\documentclass{article}
\usepackage{ijcai22}

\usepackage{times}
\usepackage{soul}
\usepackage{url}
\usepackage[hidelinks]{hyperref}
\usepackage[utf8]{inputenc}
\usepackage[small]{caption}
\usepackage{graphicx}
\usepackage{amsmath}
\usepackage{amsthm}
\usepackage{amssymb}
\usepackage{booktabs}
\usepackage{algorithm}
\usepackage{algorithmic}
\usepackage{xcolor}
\usepackage{bm}
\usepackage{float}
\newcommand{\eg}{\text{e.g.}}
\newcommand{\ie}{\text{i.e.}}

\title{Image Captioning In the Transformer Age}

\author{
Yang Xu$^1$
\and
{Li Li}$^2$
\and
{Haiyang Xu}$^3$
\and
{Songfang Huang}$^3$
\and
{Fei Huang}$^3$
\and
{Jianfei Cai}$^4$
\affiliations
$^1$School of Computer Science and Engineering of Southeast University, China\\
$^2$Bell Honors School, Nanjing University of Posts and Telecommunications, Nanjing, China\\
$^3$Alibaba Group\\
$^4$Department of Data Science and AI, Monash University\\
\emails
xuyangseu@ieee.org,
b19010805@njupt.edu.cn,
shuofeng.xhy@alibaba-inc.com,
songfang.hsf@alibaba-inc.com,
f.huang@alibaba-inc.com,
Jianfei.Cai@monash.edu}

\begin{document}

\maketitle

\begin{abstract}
Image Captioning (IC) has achieved astonishing developments by incorporating various techniques into the CNN-RNN encoder-decoder architecture. However, since CNN and RNN do not share the basic network component, such a heterogeneous pipeline is hard to be trained end-to-end where the visual encoder will not learn anything from the caption supervision. This drawback inspires the researchers to develop a homogeneous architecture that facilitates end-to-end training, for which Transformer is the perfect one that has proven its huge potential in both vision and language domains and thus can be used as the basic component of the visual encoder and language decoder in an IC pipeline. Meantime, self-supervised learning releases the power of the Transformer architecture that a pre-trained large-scale 
one can be generalized to various tasks including IC. The success of these large-scale models seems to weaken the importance of the single IC task. However, we demonstrate that IC still has its specific significance in this age by analyzing the connections between IC with some popular self-supervised learning paradigms. Due to the page limitation, we only refer to highly important papers in this short survey and more related works can be found at \url{https://github.com/SjokerLily/awesome-image-captioning}.
\end{abstract}

\section{Introduction}
Image captioning (IC) requires an AI system to describe various aspects of a given image, which include the appearing objects, the attributes of these objects, and the relationships between these objects. Although the object, attribute, and relation classification systems are ready-made, IC is more than the combinations of them since these systems usually treat the labels as numeric ids while the words in a sentence should have semantic connections. Thus, IC is a task that lies at the intersection of vision and language. Meantime, IC has its specific status among various vision-language tasks, \eg, it has a relatively long history and its key techniques are exploited in other vision-language systems. Interestingly, some key techniques of IC inherit from machine translation since, intuitively, IC can be considered as translating an image to a sentence. The two most important inherited techniques are the encoder-decoder pipeline~\cite{sutskever2014sequence,vinyals2015show} and the attention mechanism~\cite{bahdanau2014neural,xu2015show}, which establish the prototype network of almost all the following IC models. 

However, different from machine translation whose source and target are both language sentences, which can both be dealt with RNN, IC requires to deal with cross-modal data where the visual encoder is a CNN while the language decoder is an RNN. Such an encoder-decoder architecture is heterogeneous, which makes the whole model hard to be trained end-to-end (cf Sec.~\ref{sec:heterogeneous_architecture}), and thus the gap between the visual input and the language output remains large. To narrow the gap, researchers have proposed various techniques, which are summarized into three aspects in this survey: building stronger visual encoders (cf Sec.~\ref{sec:visual_feature}), designing more advanced attention mechanisms (cf Sec.~\ref{sec:attention_mechanism}), and applying vision and language structures (cf Sec.~\ref{sec:visual_language_structure}). 

Although these techniques largely improve the performances of the heterogeneous architecture, the gap is hard to be further narrowed unless we can train the whole architecture end-to-end. This inspires us to develop a homogeneous architecture to facilitate end-to-end training. Fortunately, a homogeneous architecture is ready to come out due to the appearance of Transformer~\cite{vaswani2017attention}, which has achieved astonishing success in both language and vision domains. A straightforward homogeneous architecture for IC can simply be applying Transformer as both the visual encoder and the language decoder (cf Sec.~\ref{sec:transformer_architecture}). Given this homogeneous prototype, previous experiences, \eg, the abovementioned three research directions, can be extended and incorporated to help improve the performances (cf Sec.~\ref{sec:homo_vis_lan_stru},~\ref{sec:homo_att_mech},~\ref{sec:homo_vis_lan_stru}).

\begin{figure*}[t]
\includegraphics[width=\textwidth]{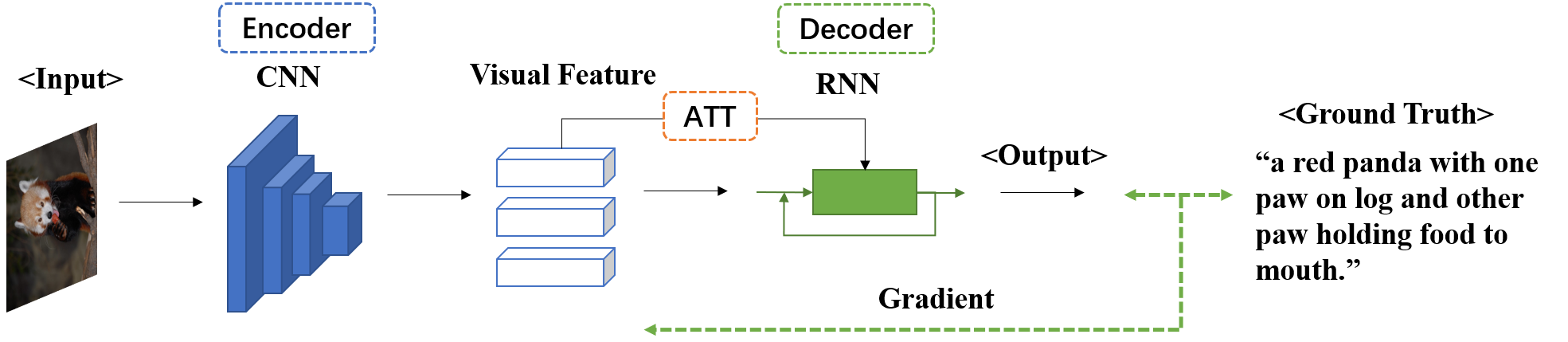}
\centering
\caption{The heterogeneous encoder-decoder architecture for IC, where the visual encoder is a CNN and the language decoder is an RNN. Since such a model is hard to be trained end-to-end, the gradient can not be backpropagated to the visual encoder.}
\label{fig:hetero}
\end{figure*}

Besides pushing the progress of individual tasks in language and vision domains, Transformer-architecture also triggered the research wave of large-scale pre-training in both domains~\cite{devlin2018bert,radford2021learning}, \ie, a large-scale model is trained by self-supervised learning and can be generalized to various downstream tasks. Similarly, the vision-language community springs up a series of self-supervised large-scale pre-training models~\cite{xu2021e2e,zhou2020unified}. Under such a circumstance, researchers may question the significance of the single IC task since it can be treated as a sub-task of these large-scale models that have stronger generalization ability. To respond to this question, we analyze the connections between IC with the large-scale models and then demonstrate that IC has its specific significance to the large-scale models (cf Sec.~\ref{sec:large_scale_pretraining}). 

To sum up, unlike some nice and exhaustive IC surveys~\cite{hossain2019comprehensive,stefanini2021show}, which detail the model architectures, the training strategies, and the quantitative comparisons by various metrics, we pay more attention to the potential future of IC architectures in this Transformer age. Another main point of this survey is that we expand the scope from IC to large-scale pre-training frameworks, where we analyze the connections between IC and some large-scale models and point out the specific significance of IC in this age.

\section{Heterogeneous Architecture}
\label{sec:heterogeneous_architecture}
Figure~\ref{fig:hetero} sketches the classic encoder-decoder architecture for IC, which contains a CNN-based visual encoder, an RNN-based language decoder, and a cross-modal attention block. In this architecture, the visual encoder extracts the visual features and these features are input into the language decoder for captioning. Since the visual encoder (CNN) and the language decoder (RNN) do not share the same structure, this architecture is considered heterogeneous. 

The major problem of such heterogeneous architecture is that the whole model is hard to be trained end-to-end. This is mainly because CNN and RNN do not share the basic network component, then the optimization strategies, \eg, the optimizer or the learning rate, of the encoder and decoder are hard to be unified. To remedy this, researchers divide the training of the visual encoder and the language decoder. Specifically, they pre-train a visual encoder and then fix it. When the model is trained by the caption supervisions, the parameters of the visual encoder will not be updated. As a result, the gradients are not backpropagated from the word-level supervision to the pixel-level input, as shown in Figure~\ref{fig:hetero} that the gradient (the green dash line) does not transmit to the CNN. This means that these heterogeneous architectures are not really end-to-end trained. Thus, the visual encoder fails to learn high-level semantic knowledge from the caption supervisions and the extracted visual features have determined the upper bound of the generated captions' quality, where the gap between vision and language domains is still huge. To narrow the gap, various techniques have been proposed to refine this heterogeneous architecture. For simplicity, we cluster them into three main directions in the following subsections.

\subsection{Visual Feature}
\label{sec:visual_feature}
One direction is to improve the quality of the visual encoder for extracting better features, \ie, training more sophisticated architectures by more images with more semantic annotations. At the beginning, ~\cite{vinyals2015show} set the visual encoder as a GoogLeNet to extract the feature of the whole image as the input to the decoder. However, such image-level features may miss the details of some specific regions. To address it, ~\cite{xu2015show} and~\cite{anderson2018bottom} respectively divided an image into uniform gird regions and salient object regions and then introduced the attention mechanism to selectively describe these regions. In this way, the generated captions can be more grounded to the image that more details of the image can be described. Besides the evolution from image-level to region-level features, the pre-training strategy has also been developed. At the early stage, only image classification was used~\cite{vinyals2015show,xu2015show}, and then object detection and attribute classification were respectively introduced by~\cite{lu2018neural} and~\cite{anderson2018bottom} to train a ResNet-based Faster RCNN for generating more descriptive captions. 

\subsection{Attention Mechanism}
\label{sec:attention_mechanism}
Another direction, maybe the most popular one, is to develop more advanced attention mechanisms. Intuitively, the attention mechanism works as a bridge to narrow the gap between vision and language domains, \ie, it makes the decoder dynamically describe the selected visual regions conditioned on the partially generated sentence during captioning. 

Since~\cite{xu2015show} introduced the attention mechanism in IC, researchers developed such mechanism from two aspects. The first one is the attended source, \ie, what contents should be attended. For example, ~\cite{xu2015show,jiang2020defense} and~\cite{anderson2018bottom,lu2018neural} proposed to attend to uniform grid regions and salient object regions, respectively. Some researchers also showed that directly attending to semantic tags like categories or attributes~\cite{you2016image,yao2017boosting} is also helpful. Besides, features from multi-channels~\cite{chen2017sca} or multi-CNNs~\cite{jiang2018recurrent} were also used.

The second aspect is to modify the architecture of the attention module, \ie, how to compute the attention weights. Different from the widely used inner-product attention (Eq.6 of~\cite{bahdanau2014neural}) in the NLP field, additive attention (Eq.2 of~\cite{xu2015show}) is more popular in IC. This is mainly because the visual encoder and language decoder are heterogeneous, which makes the extracted visual and word representations do not stay in the same representation space. Thus using inner-product to calculate the similarity is not reasonable. In contrast, the additive attention applies two learnable matrices to respectively map the visual and word representations into similar representation spaces for getting the similarity, which is more suitable in the IC case. Such additive attention module was also stacked~\cite{ke2019reflective,wang2020show} to build more sophisticated language decoders for better captions.

\subsection{Visual and Language Structure}
\label{sec:visual_language_structure}
Another mainstream direction is to incorporate the visual and language structures into the encoder-decoder framework. In the middle of the pure-vision and vision-language tasks, the scene graph is a transitional data structure that is widely used as the mediator to narrow the vision-language gap. The scene graph contains three kinds of nodes: objects, attributes, and relations. The objects are connected through their pairwise relations and the attributes are connected to the corresponding objects. Such a data structure contains useful semantic and topological knowledge, which can facilitate the IC models to generate more descriptive captions. To incorporate scene graphs,~\cite{yao2018exploring,yang2020auto} deployed Graph Neural Networks~\cite{battaglia2018relational} to embed them and input the embeddings to the language decoder for captioning.~\cite{chen2020say} further proposed to adjust the scene graphs for controllable captioning. Besides scene graphs, the tree-based encoder was also proposed in~\cite{yao2019hierarchy} to learn the hierarchical structures of images.

In addition to the visual structure, the text pattern was also exploited for designing advanced IC models. ~\cite{yang2021auto} observed that the sentences can be parsed as trees and then designed a tree-based encoder-decoder for learning the hidden tree structures. Some other researchers acted in an opposite way that they divided the integral sentence into distinguishable kinds of words based on the sentence pattern and then designed diverse module networks to generate them. For example, after observing that some words are related to the visual contents and some are not,~\cite{lu2017knowing} designed two modules for vision and non-vision words, respectively. ~\cite{yang2019learning} went one step further by designing four different modules, which are object, attribute, relation, and function modules, for captioning.

\begin{figure*}[t]
\includegraphics[width=\textwidth]{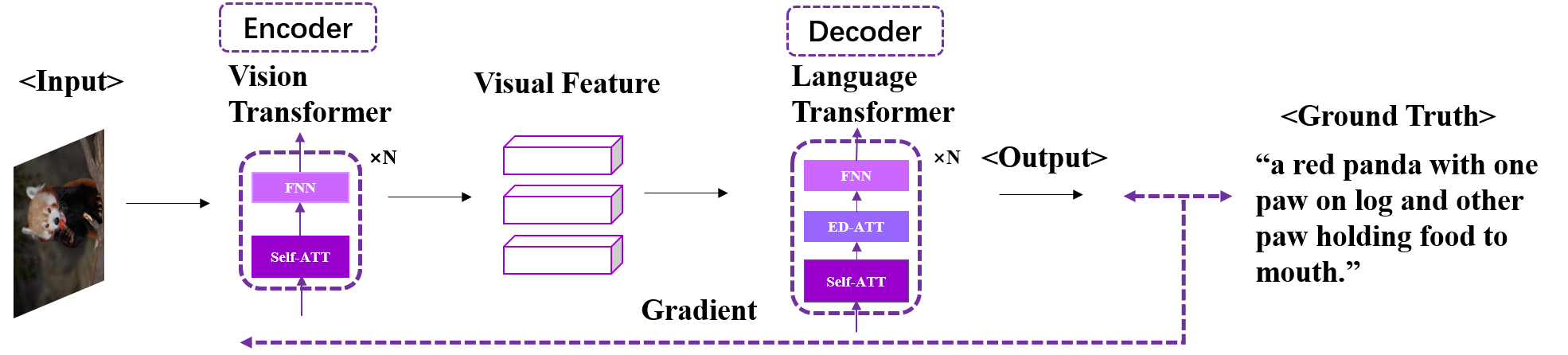}
\centering
\caption{A Transformer-based homogeneous encoder-decoder architecture for IC, where the visual encoder and language decoder are both Transformer networks. Since this homogeneous architecture facilitates the end-to-end training, the gradient can be backpropagated to the visual encoder.}
\label{fig:homo}
\end{figure*}

\subsection{Limitations}
No matter how the heterogeneous architectures are improved, their visual encoders are fixed after pre-training and thus fail to learn anything new from the caption supervisions. Moreover, the annotations (categories or attributes) used to pre-train these encoders are represented by numeric ids instead of the language words. Thus the extracted features may also contain these numeric ids instead of the language knowledge. When the whole model is trained by the captions, the decoder will first recognize these ids from the visual features and then connect them with the words, \eg, recognizing the id ``APPLE'' from the image with apples and connecting this id to the word ``apple'' or ``apples''. While in this process, spurious correlations~\cite{yang2021deconfounded} may be created and the dataset bias will be learned~\cite{hendricks2018women}. All these drawbacks motivate us to design a homogeneous architecture that facilitates end-to-end training, which can be achieved by exploiting the Transformer architecture, as detailed in the next section.
 
\section{Homogeneous Architecture}
\subsection{Transformer}
\label{sec:transformer_architecture}
The Transformer architecture was firstly proposed in~\cite{vaswani2017attention} to address machine translation. The major technical component in this architecture is the multi-head attention block. Specifically, given three input sequences: the query $\bm{Q}=\{\bm{q}_1,...,\bm{q}_{d_Q}\} \in \mathbb{R}^{d_Q \times d}$, the key $\bm{K}=\{\bm{k}_1,...,\bm{k}_{d_K}\} \in \mathbb{R}^{d_K \times d}$, and the value $\bm{V}=\{\bm{v}_1,...,\bm{v}_{d_V}\} \in \mathbb{R}^{d_V \times d}$, this block calculates the output $\bm{O}$ as follows:
\begin{equation} \label{equ:multi-head}
\small
\begin{aligned}
 \textbf{Input:} \quad  &\bm{Q},\bm{K},\bm{V} \\
 \textbf{Att:} \quad  &\bm{A}_i=\text{Softmax}( \frac{\bm{Q}\bm{W}_i^Q(\bm{K}\bm{W}_i^K)^T}{\sqrt{d}} ) \\
 \textbf{Head}:  \quad  &\bm{H}_i=\bm{A}_i\bm{V}\bm{W}_i^V,\\
 \textbf{Multi-Head:} \quad & \bm{H}= [\bm{H}_1,\bm{H}_2,...,\bm{H}_8]\bm{W}^H, \\
 \textbf{Output:} \quad  &\bm{O}=\text{FFN}(\bm{H}), \\
\end{aligned}
\end{equation}
where $\bm{W}_i^Q, \bm{W}_i^K, \bm{W}_i^V \in \mathbb{R}^{d \times d_h}$, and $\bm{W}_i^H \in \mathbb{R}^{d \times d}$ are all trainable matrices; the number of the heads is set to $h$ and $d_h=d/h$; $\bm{H}_i$ is the output of the $i$-th head and $\bm{A}_i$ is the corresponding attention matrix; and FFN is Position-wise Feed-Forward Network, whose structure is FC-RELU-FC.

By the second step ``Att'' in Eq.~\eqref{equ:multi-head}, the inner-product of each $\bm{q}_i$ and $\bm{k}_j$ is calculated, thus it can learn to capture the dense correlations between the query and key sequences. Interestingly, when setting $\bm{Q}=\bm{K}=\bm{V}$, this attention mechanism is called self-attention, which can capture the long-range dependencies between the elements at the same sequence. When $\bm{K},\bm{V}$ and $\bm{Q}$ respectively come from the encoder (the source domain) and the decoder (the target domain), such mechanism is called encoder-decoder attention (or the cross-modal attention), which can capture the dense correlations between the source and target domains. A classic Transformer-architecture contains both of the two attention mechanisms, and thus it can learn dense correlations among one sequence or between different sequences. Exactly because the word correlations play crucial roles in solving various NLP tasks, Transformer has achieved astonishing success in this field. Furthermore, Transformer also demonstrated huge potential in pure-vision domains and numerous Transformer-based architectures have been proposed to address different vision tasks~\cite{khan2021transformers}. 

Motivated by such progress, a pure Transformer-based homogeneous encoder-decoder captioner is ready to come out. As sketched in Figure~\ref{fig:homo}, a straightforward homogeneous architecture can be configured as follows: the visual encoder is set as a pre-trained vision Transformer~\cite{liu2021swin} and the language decoder is set as a classic Transformer~\cite{vaswani2017attention}. Since now the architecture is homogeneous that the optimization strategies of both the encoder and decoder can be unified, the whole model can be end-to-end trained, \ie, as shown in Figure~\ref{fig:homo}, the gradient (the purple dash line) can be backpropagated from the word-level supervision to the visual Transformer. In this way, the visual encoder can learn high-level semantic knowledge from the language supervisions, while the encoder of the previous heterogeneous architecture can not. Given this homogeneous prototype, researchers can further refine it for generating better captions. To help the readers get some preliminary ideas, we now reexamine the three directions mentioned in Section~\ref{sec:heterogeneous_architecture} here. 

\subsection{Visual Feature} 
\label{sec:homo_visual_feature}
In the heterogeneous architecture, using more semantic annotations to pre-train a visual encoder could remedy the information loss caused by the divided training. A homogeneous architecture encourages the end-to-end training and thus the visual encoder could learn more high-level semantic knowledge. While such gain does not come free in the sense that training a multi-modal Transformer architecture may require much more image-text pairs and GPU resources. Thus, if the training data or GPU resources are not sufficient, a ready-made pre-trained vision Transformer is still necessary to provide a good warm-start. Researchers could follow the progress of vision Transformer~\cite{khan2021transformers} and draw lessons about training strategies from some classic captioning research like~\cite{anderson2018bottom,lu2018neural} to build stronger visual encoders.

\subsection{Attention Mechanism}
\label{sec:homo_att_mech}
Transformer is one architecture that is almost full of the stacked attention operations, \ie, self-attention and encoder-decoder attention. Both operations come from Eq.~\eqref{equ:multi-head} by changing the sources of $\bm{Q},\bm{K},\bm{V}$. Thus, researchers can follow this strategy to change the sources of $\bm{Q},\bm{K},\bm{V}$ in a captioning model to construct new attention blocks, \eg, using semantic tags, multi-level or multi-CNN features, and an augmented memory~\cite{cornia2020meshed}.

As discussed in Section~\ref{sec:attention_mechanism}, besides changing the attention sources, more complex architectures of the attention operations can be designed. Since the encoder and decoder are homogeneous and the encoder-decoder attention is the inner-product operation, the visual features and word embeddings will be encouraged to stay in similar representation spaces. Thus some of the attention mechanisms used in the heterogeneous architecture may not be powerful in this homogeneous architecture. Interestingly, some researchers designed certain more complex attention operations based on Eq.~\eqref{equ:multi-head}, \eg, AoA~\cite{huang2019attention} or X-LAN~\cite{pan2020x}, and achieved higher performances than the corresponding baselines. Although these operations were originally deployed in the heterogeneous architecture, researchers can still try to replace Eq.~\eqref{equ:multi-head} by them or some newly designed ones. 

\subsection{Visual and Language Structure}
\label{sec:homo_vis_lan_stru}
From the perspective of the graph neural network, self-attention is one kind of fully-connected graph operation~\cite{battaglia2018relational} that calculates the attention weight between every two elements of one input sequence. The calculated attention weight can be treated as the corresponding edge weight. However, the hidden structure of vision and language data may not be the fully-connected graphs, but more likely to be sparse graphs or hierarchical trees~\cite{yao2019hierarchy,yang2021auto}. Thus one good research direction is to figure out how to incorporate the sparse and hierarchical inductive bias into the typical Transformer.

Some vision and language models have respectively introduced different hierarchical inductive bias into the Transformer architectures and achieved improvements. For example, Swin-Transformer~\cite{liu2021swin} only calculated self-attention in one swift window and considered the multi-scale characteristic of the visual features. Tree-Transformer~\cite{wang2019tree} enabled the model to automatically parse a dependency tree from the input sentence. Motivated by their success, researchers can simply set the encoder/decoder as these ameliorated vision/language Transformers. Furthermore, since the hierarchical structures of the vision and language have different expression forms, \ie, images are constructed by natural vision scenes while sentences are constrained by the grammar rules, researchers should consider more about how to align or transfer the diverse structures between two domains.  

As discussed in Section~\ref{sec:heterogeneous_architecture}, a heterogeneous structure is hard to be end-to-end trained. Incorporating more complex components for learning hierarchical structures will further aggravate the training difficulty. Thus exploiting hierarchical inductive bias in a heterogeneous structure may not totally release the power of such a strategy. For example,~\cite{yang2020auto} discovered that when using the scene graphs to generate the captions, the captions generated from language scene graphs (177.0 CIDEr) largely outperform the ones generated from visual scene graphs (129.6 CIDEr). Such a huge gap also demonstrates the enormous potential of further transferring the hierarchical structures between two domains.

Besides transferring the whole hierarchical structures from the language domain to the vision domain, researchers also try to decompose such structure into various sub-tasks and then design diverse modules to solve them~\cite{yang2019learning,lu2017knowing}. However, since these module networks use the pre-trained visual features as the inputs, then if the visual features have suffered from information loss, each module may fail to learn what they should learn. For example, if the visual encoder is pre-trained by object detection, the extracted visual features may contain less information about object attributes and thus the attribute module may fail to generate the corresponding attribute words.

In the homogeneous architecture, we can set each module network and the module controller as a Transformer. Among them, the module networks are designed to provide different visual patterns or the language context and the module controller is designed to control which module networks should be used when generating a specific word. To further encourage each module to learn its corresponding knowledge, the part-of-speech of the words in the ground truth captions can be used as additional supervision. Since all these modules and the controller are homogeneous Transformer networks, the whole model can be well-trained. As a result, each module network can learn the corresponding knowledge and suffer less information loss than the heterogeneous counterpart. 

\section{In the Age of Large-scale Pre-Training}
\label{sec:large_scale_pretraining}
As discussed in the survey of Prompt-based Learning~\cite{liu2021pre}, NLP fields have encountered two sea changes in recent years. In the first one, fully-supervised learning is replaced by the ``pre-train and fine-tune'' paradigm~\cite{devlin2018bert}. And since 2019, more and more research works~\cite{radford2019language,petroni2019language}  follow a novel learning paradigm, which is the second sea change: ``pre-train, prompt, and predict''. Both of them pre-train a large-scale model on the dataset with a huge amount of texts by self-supervised learning and the pre-trained model can be easily generalized to various downstream tasks.

Motivated by these two sea changes in the NLP field, researchers in the vision-language domain also propose to train the large-scale Transformer architectures on millions of web-collected image-text pairs by self-supervised learning~\cite{zhou2020unified,xu2021e2e}. Some of these multi-modal large-scale pre-training models can also be used for IC and have achieved much better performances than small-scale ones~\cite{li2020oscar}. Observing these achievements, we may ask: Shall we still focus on small-scale IC systems? Our answer is affirmative since IC can contribute its technical reserves to these large-scale models. Moreover, more and more researchers show that captioning data can help solve pure vision tasks. All of these demonstrate that IC has its specific time significance in this large-scale pre-training age. 

\subsection{Pre-training and Fine-tuning}
Transformer architecture assumes less inductive bias compared with CNN and LSTM~\cite{khan2021transformers}, and thus it is more suitable to be trained by self-supervised pretext tasks on large-scale data, \eg, millions or billions of image-text pairs. However, not every research group can afford such huge training burdens. It is therefore urgent to develop more economic Transformer-based large-scale multi-modal models by incorporating more inductive bias about vision and language data. However, most of the existent large-scale models focus on designing new training objectives or adjusting the integral training strategy. For example,~\cite{chen2019uniter} proposed some new training objectives like Masked Region Classification, Masked Region Feature Regression, and Masked Region Classification to improve the traditional Masked Region Modeling. While~\cite{li2021align} adjusted the training strategy from ``fusing before aligning'' to ``aligning before fusing'' to learn better image-text connections.

However, few of them consider the hidden structures of the vision and language data and also do not incorporate the corresponding inductive bias into the model, while incorporating such inductive bias could improve the data utilization efficiency and decreases the amount of training data. Interestingly, considering how to incorporate such inductive bias is one main research direction in the IC field and these ideas could provide meaningful prototypical enlightenment for designing more economic multi-modal large-scale models. Furthermore, since training captioning models would be more economic than large-scale models, researchers could afford the trial-and-error procedure. Thus, the progress of IC can also push forward the research of large-scale models.

\subsection{Pre-training and Prompt}
As introduced in~\cite{liu2021pre}, different from the traditional supervised learning, which learns a specific model $p(\bm{y}|\bm{x})$ to predict the specific target $\bm{y}$ from the input $\bm{x}$, prompt-based learning unifies various NLP tasks as the single one: language modeling. For a specific task, given the input text $\bm{x}$, it is firstly re-formulated to a new texture string prompt $\bm{x}'$ with some unfilled slots. Then the language model will fill these slots based on the contextual knowledge to get a completed string $\hat{x}$, from which the final output $\bm{y}$ will be derived. Take machine translation as one simple example, to translate one English sentence ``there is a dog.'' to the Chinese one, we can create a prompt string: ``English: there is a dog. Chinese: [\textsc{slot}].'' and input this prompt string into a well-trained language model to fill in the [\textsc{slot}] token. 

By this learning paradigm, to solve a great amount of diverse NLP tasks, we only need one well-trained language model instead of training different models on various datasets. In this process, the only thing we need to do is to design the suitable ``prompt engineering'', \ie, re-formulating the text inputs of different tasks as suitable prompt templates. By prompt engineering, the large-scale model can also be pre-trained by various NLP tasks and thus can learn the representations with stronger generalization ability compared with the models trained by one single task.

Analogically, this learning paradigm is introduced in the vision-language field that various tasks are unified as the language generation conditioned on the multi-modal contexts, \ie, the image and the prefix words~\cite{cho2021unifying,tsimpoukelli2021multimodal,wang2021simvlm}. For example,~\cite{cho2021unifying} unified various vision-language tasks, \eg, Visual Question Answering, Visual Reasoning, Referring Expression Comprehension, Visual Commonsense Reasoning, Image Captioning, and Multi-modal Machine Translation, into one framework. Here we use Visual Question Answering (VQA) as the example to show the differences between this paradigm with the traditional one. In a traditional VQA model, given an image and a question, the visual/language encoders are respectively used to embed them. Then a cross-modal attention block will take the visual/language embeddings as the input to learn the image-question connections for the final prediction. However, this prediction problem is usually treated as the classification~\cite{antol2015vqa} that each answer is labelled as a numeric id, which loses the language semantic of the answer. While in the prompt-based framework, the input is re-formulated as: ``[Image Patches] [VQA] [Question] [Answer]''. Here each `[]' denotes one specific token: [Image patches] denotes the divided patches of an image, [VQA] is a specific learnable embedding used to tell the model this is a VQA task, [Question] is the given question, and [Answer] is a slot which needs to be filled in. Since the unified model is a language generator, the [Answer] slot will be filled in an answer sentence, which is more according to the characteristic of VQA than the traditional ones.

Actually, the task of generating texts conditioned on multi-modal inputs is what an IC model is doing. Specifically, when an IC model generates a sentence, it generates the word one-by-one conditioned on the given image and the context of the partially generated caption. Thus, although the scale of training an IC model is smaller than a prompt-based vision-language model, the nature of the two models are the same. In this way, all the fruits of captioning research can construct a solid cornerstone for further improving these prompt-based models, \eg, more advanced attention mechanisms and better structure priors. Besides them, the training objective (\eg, the reinforcement learning-based strategy~\cite{rennie2017self}) or even the metric (\eg, CIDEr~\cite{vedantam2015cider}) of the captioning systems can provide useful insights to train or measure these prompt-based models. 

\subsection{Solving Pure Vision Tasks}
Besides these vision-language large-scale models, researchers go one step further by exploiting captioning data to solve pure vision tasks like classification~\cite{radford2021learning}, detection~\cite{zareian2021open}, or even image generation~\cite{ramesh2021zero}. The traditional pure-vision datasets usually have the following drawbacks: 1) each task requires one specific kind of annotations for the same image (\eg, object and attribute classifications require two types of annotations); 2) they only contain a fixed set of predetermined labels; 3) the labels are treated as numeric ids instead of the semantic words. Compared with these datasets, captioning data provide a more comprehensive description of the various aspects of an image, \eg, the object categories, attributes, or relations. More importantly, these annotations are semantic words and have a much larger label space for learning more semantic visual representations.

However, captioning data mixes up the different concepts, which largely increases the learning difficulty of a model. One way to address this challenge is to feed the model with huge amounts of image-text pairs. For example, CLIP~\cite{radford2021learning} collected 400 million image-text pairs from the internet and applied contrastive learning to train a large-scale Transformer-based architecture. By exploiting the prompt paradigm, the trained model can successfully achieve zero-shot transfer to various downstream vision tasks, which demonstrates the potential of using image-text data to solve various vision tasks. However, the efficiency is still a big challenge since it requires 18 days and 592 V100 GPUs to train CLIP. 

Although the calculation of contrastive loss, \ie, checking whether an image is aligned with a text, is faster than the captioning loss, \ie, generating the words one-by-one, the data utilization efficiency is also decreased. Thus researchers may still use the captioning loss while meantime designing more efficient architectures to lower the training burden. For example, the modular network can disentangle the mixed up concepts by various modules for addressing different concepts. The experiments in~\cite{yang2019learning} show that the modular architecture requires fewer data to achieve the same performances compared with non-modular architectures. Also, other structure inductive bias (cf.Sec~\ref{sec:visual_language_structure}) can be incorporated into the model to improve the data utilization efficiency.

\section{Conclusion}
In this survey, we first analyzed the major drawback of the classic heterogeneous IC architecture that the whole model is hard to be trained end-to-end.
Then we briefly reviewed three major strategies for alleviating this drawback, which include constructing better visual encoders, designing more advanced attention mechanisms, and incorporating more structural inductive bias. However, these models are still hard to be trained end-to-end. To solve this natural defect, we discussed the feasibility of building a Transformer-based homogeneous architecture for facilitating the end-to-end training. Meantime, we showed some preliminary ideas from the heterogeneous architectures for improving the homogeneous one. Lastly, we analyzed the connections between IC with the Transformer-based large-scale pre-training paradigm, which is one of the most popular research directions today. Through the analysis, we demonstrated that although the appearance of these large-scale models may distract the researchers' attention from IC, the nature of the research contents does not change so much. Thus, IC still has its specific significance in this Transformer age.

\newpage
\bibliographystyle{named}
\bibliography{ijcai22}

\end{document}